# A single target tracking algorithm based on Generative Adversarial Networks

Zhaofu Diao[1]    Ying Wei[1]    Yujiang Fu[1]    Shuo Feng[1]

[1]Northeastern University, China

## Abstract

*In the single target tracking field, occlusion leads to the loss of tracking targets is a ubiquitous and arduous problem. To solve this problem, we propose a single target tracking algorithm with anti-occlusion capability. The main content of our algorithm is to use the Region Proposal Network to obtain the tracked target and potential interferences, and use the occlusion awareness module to judge whether the interfering object occludes the target. If no occlusion occurs, continue tracking. If occlusion occurs, the prediction module is started, and the motion trajectory of the target in subsequent frames is predicted according to the motion trajectory before occlusion. The result obtained by the prediction module is used to replace the target position feature obtained by the original tracking algorithm. So we solve the problem that the occlusion causes the tracking algorithm to lose the target. In actual performance, our algorithm can successfully track the target in the occluded dataset. On the VOT2018 dataset, our algorithm has an EAO of 0.421, an Accuracy of 0.67, and a Robustness of 0.186. Compared with SiamRPN ++, they increased by 1.69%, 11.67% and 9.3%, respectively.*

1. Introduction

Single-target tracking has always been a hot trend in the field of computer vision [1,2,3,4,5], and people have been improving the performance of existing algorithms [6,7,8,9,10,11,12,13]. The current tracking algorithms can be roughly divided into two categories: one is traditional method based on Correlation Filter [14,15,16,17], the correlation filtering between the template frame and the detected frame after Fourier transform is used to obtain the target of the subsequent frame. The second is a deep learning method based on Convolutional Neural Network (CNN) [5,18,19,20], which convolves the image features of the template frame and the detected frame to obtain the tracking target. With the rapid development of deep learning, CNN-based feature representation has been adopted by more and more people. Because of its powerful feature expression ability, it has performed well in the fields of image classification, target detection, and pedestrian recognition.

Single target tracking is often treated as a problem about the similarity between template frames and detection frames. People attempt to continuously track the targets in subsequent frames through a given region of interest. However, the subsequent frames may have external factors such as occlusion, deformation, and illumination changes [5,21]. At this point, the response of the detected frame will change drastically, causing tracking deviation. Therefore, one of the main topics in the single target tracking field is to design an algorithm that can cope with the above problems and is robust as much as possible.

In order to solve the problem of target deformation and illumination change on the detection frame, there are currently two workarounds, namely online update templates and fixed templates. The strategy of updating the template frame online can improve the robustness of the model to deformation and other problems; the fixed template frame does not change the feature of the template frame, and can better cope with the situation when the target disappears for a long time and then returns to the field of vision. Whether updating templates or fixing templates, the key to the problem lies in the distinction between foreground and background, where the background contains both a semantic background and a non-semantic background.

In order to solve the above problems, we propose a tracking framework. In addition to the Region Proposal Network (RPN) for extracting image features, we have added an occlusion awareness module and a path prediction module. The occlusion awareness module determines whether the target in the detection frame is occluded by other objects through the heat map and classification branch score. The path prediction module is activated when the target is occluded, and uses the Generative Adversarial Network (GAN) to predict the trajectory of the target that is most likely to be taken in the next few frames. At the same time, we found that if the object is occluded during the training, the information learned by the model is mostly interference information. Therefore, we use the sample when the object is occluded as a negative sample during training, which makes the model learning more accurate.

The algorithm is mainly composed of the RPN that acquires the target according to the image feature and the GAN that acquires the target according to the location feature. Our solution mainly solves the problem of tracking failure caused by occlusion in the field of single target tracking. Our method was evaluated on the VOT2016 and VOT2018 datasets and the results were very competitive.

2. Related work

Before the success of the CNN network, people had turned their attention to the method based on correlation filter [22] and the traditional target tracking algorithm. Later, the Siamese Network attracted people's attention due to its superior performance [19,23,24,25]. In 2015, a tracking method based on Siamese Network (SiamFC) [25] attracted people's attention. The algorithm combines the idea of deep learning with correlation filter, and transforms the target tracking problem into the similarity matching problem. The algorithm achieved good results. However, with the deepening of the research, people find that this method has problems: First, the highest score in the response diagram may not be the tracking target, but other similar objects. which may interfere with the result. Second, this method is only applicable to the case where the scale change of the tracking target is small. For objects with large scale changes, the performance of the algorithm will decrease. Third, tracking will fail when occlusion is encountered.

On this basis, people introduced the RPN from the detection algorithm and proposed an improved method to convert the tracking problem into a combination of classification and regression. SiamRPN [26] was born. SiamRPN solves the problem that SiamFC can't automatically adapt to the scale during the tracking process. In 2018, DaSiamRPN [27] was proposed to solve the problem of intra-class interference, and also solved the problem of how to continue tracking when the target reappears after a short-term disappearance.

SiamRPN++ [28] is a further improvement of the siamese network, which mainly solves the problem that the depth of the network is limited and the available deep semantic features are few. SiamRPN++ improved the network and removed the padding. Then, the feature maps of the 3rd, 4th, and 5th layers are input into the RPN, and finally the results of classification and regression are obtained. In the same period, there is another algorithm that uses ResNet's idea to deepen and widen the network structure [29].

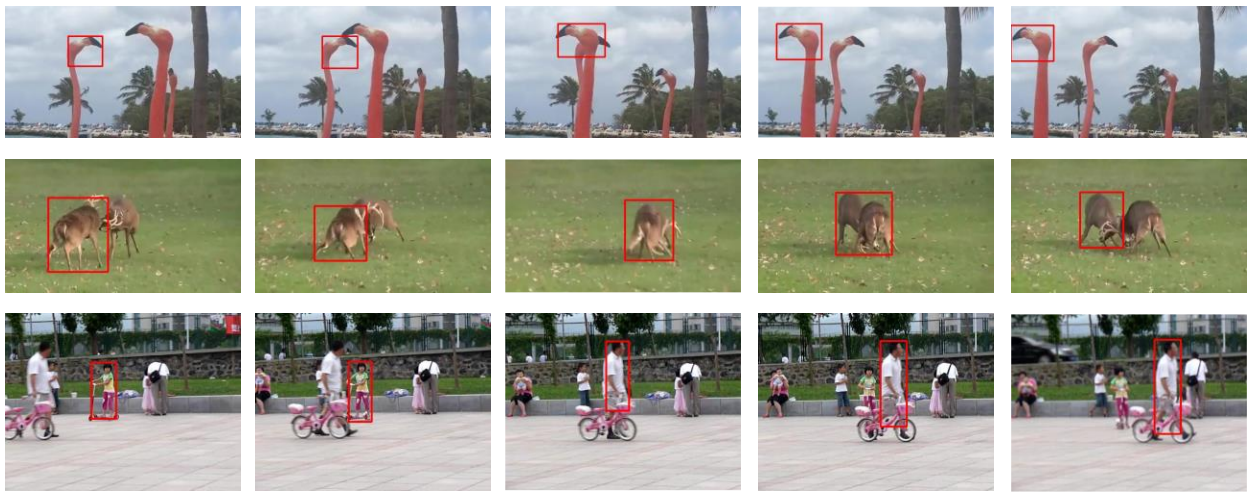

Figure 1. Occlusion problem leads to wrong target

So far, all algorithms do not solve the problem of the target being occluded. As shown in Figure 1, occlusion may result in loss of target, so it is difficult to adopt the method based on siamese networks in actual products. People are more willing to use the target re-identification algorithm to solve the occlusion problem in tracking. However, the premise is that at the moment of occlusion, other cameras must be able to capture unobstructed targets. Which requires more than two cameras per target, and the number of cameras needs to be doubled at least. Therefore, an inexpensive method for solving the target occlusion problem needs to be proposed.

3. Anti-occlusion tracking algorithm

In order to solve the above problem, we propose a single target tracking algorithm with anti-occlusion capability, and name it Siam-GAN, which closely combines target tracking with path prediction. Our method is mainly composed of the RPN that acquires the target according to the image feature, the occlusion awareness module, and the GAN that acquires the target according to the location feature. We use the RPN to obtain tracking targets and potential interferers, and use the occlusion awareness module to determine whether the interferers are occluding the target. If no occlusion occurs, tracking continues. If occlusion occurs, the prediction module is activated, and the motion trajectory of the target in subsequent frames is predicted according to the motion trajectory before occlusion. The result obtained by the prediction module is used to replace the target's position feature obtained by the original tracking algorithm to solve the problem that the occlusion causes the tracking algorithm to lose the target. Our solution mainly solves the problem of tracking failure caused by the occlusion of the target in the single target tracking field. Our algorithm performs better than some current tracking algorithms.

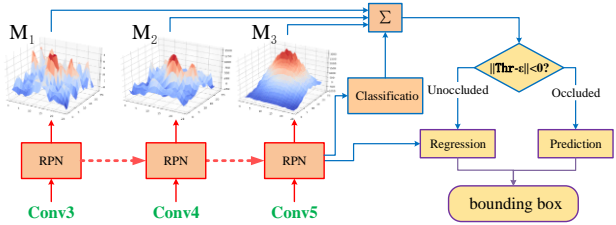

Figure 2. Occlusion awareness module. The occlusion awareness module consists of two parts: the first part is to obtain three heat maps according to the three convolutional layers of the siamese network, and calculate the distance between the target and the interference object on the three heat maps to determine whether occlusion occurs. The second part is to determine whether it is occlusion based on the classification layer score of the RPN module corresponding to the fifth convolutional layer.

3.1. Occlusion awareness module

We have designed an occlusion awareness module. The module consists of two parts: The first part is to obtain the three different levels of feature maps from the 3rd, 4th, and 5th convolutional layers of the siamese network. Enter them into the RPN module separately to get three heat maps. Then calculating the distance between the target and the interference object on the three heat maps, and summing the distance according to the weight. If the distance is less than the threshold set by the algorithm, it is determined to be occlusion. The second part is judged by the score of the classification layer, which is obtained after inputting the output of the fifth convolutional layer to the RPN module. If the score is less than the threshold set by the algorithm, it is determined to be occlusion. We will sum these two parts according to the weights and use them as the final occlusion judgment standard.

After the training of the unobstructed supervised siamese network, we analyzed the feature of the three different levels from the 3rd, 4th, and 5th convolutional layers of the network. We found that when occlusion occurs, the shallow features fluctuated significantly, the fluctuations in the middle layer are stable and the fluctuations in the deep layer are small. We combine the heat map with the classification branch of the siamese network, and propose an occlusion awareness module that can accurately determine whether the target is occluded, as shown in Figure 2.

In the response feature graph, the peaks in the neighborhood of the highest peak are still higher than other peaks. So this problem can't simply get the highest k peaks. It is also necessary to eliminate the coordinates in each neighborhood in order to obtain the exact coordinates of each peak position. We use the greedy algorithm to deal with this problem, retaining the eligible conditions and rejecting non-conforming conditions.

We express the characteristic fluctuation matrices from the 3rd, 4th, and 5th convolutional layers of the network as:
$$M_i(n,n) = \psi_i(z) \times \psi_i(z) \quad (1)$$
Where $i$=1, 2, 3.

In each matrix, we believe that the highest peak of the response is the target we want to track, and other peaks may be interferers. In order to determine whether the interferer has an influence on the target, it is necessary to analyze the distribution of the highest peak and other peaks. We first obtain the highest k peaks in each characteristic fluctuation matrix, and obtain the peak coordinates $(x, y) = (x_i^j, y_i^j)$ from the three characteristic matrices. In the formula, $i = 1, 2, 3$; $k_i$ represents the number of peaks in the characteristic fluctuation matrix, $j \in (1, k_i)$. Then we calculate the Euclidean distance between each peak and the highest peak in the three matrices:
$$dis_i = \sum_{j=2}^{k_i} \sqrt[2]{\left(x_i^1 - x_i^j\right)^2 + \left(y_i^1 - y_i^j\right)^2} \quad (2)$$
If the distance $dis_i$ is smaller than the threshold we set, it can be judged that the interference object obscures the target. In order to make the judgment more precise, we will sum the peak distances at each level by weight:
$$Dis = \sum_{i=1}^{3} \eta_i \min(dis_i) \quad (3)$$
We set the threshold $d$ for judging occlusion. When $Dis < d$, we judge that the target is occluded.

In order to prevent the occlusion awareness module from erroneously judging the normal tracking condition as occlusion, we use the positive sample score of the classification branch to supervise. This score is obtained by inputting the feature map from the fifth convolutional layer to the RPN module. This score will also have a certain weight in the judgment. We take the peak distance and classification score into account to get the final occlusion judgment standard.

The pseudo code of the algorithm is as follows, there are 6 functions, GET_TOP is the function of finding the peak, the other functions are the post-processing of the peak, and finally the integrated peak distance is obtained by COMPUTE_DIS.

**Algorithm 1** Compute Distances among peaks
**Input**: n×n
**Output**: distance
```
 1:  function GET_TOP(Array,k)
 2:      ori_index ← []
 3:      for each ind ∈ topk_index do
 4:          row ← ind // n
 5:          col ← ind % n
 6:          ori_index ← ori_index + (row,col)
 7:      end for
 8:      return ori_index
 9:  end function
10:  function REMOVE_LOW_POINT(topk,score)
11:      top1 ← topk[0]
12:      top1_poi ← score[top1[0]top1[1]]
13:      new_top4 ← []
14:      new_top4 ← new_top4 + top1
15:      for each iq ∈ [1,length(topk)] do
16:          if score[topk[iq[0]]topk[iq[1]]] > 0.75 × top1_poi then
17:              new_top4 ← new_top4 + topk[iq]
18:          end if
19:      end for
20:      return new_top4
21:  end function
22:  function IS_NEIGHBOR(ori_pos,insert_pos)
23:      flag1 ← insert_pos[0] - oripos[0] + 2
24:      flag2 ← insert_pos[0] - oripos[0] - 2
25:      flag3 ← insert_pos[1] - oripos[1] + 2
26:      flag4 ← insert_pos[1] - oripos[1] - 2
27:      if flag1<=0 && flag2>=0 && flag3<=0 && flag4>=0 then
28:          return True
29:      end if
30:      return False
31:  end function
32:  function CHECK_NEIGHBOR(ori_arr,insert_pos)
33:      for each arr_ind ∈ [0,length(ori_arr)] do
34:          ff_flag ← IS_NEIGHBOR(ori_index[arr_ind],insert_pos)
35:          if ff_flag == True then
36:              return True
37:          end if
38:      end for
39:      return False
40:  end function
41:  function MERGE_NEIGHBOR(arr_p)
42:      new_list ← []
43:      new_list ← new_lise + arr_p[0]
44:      for each ind ∈ [1,length(arr_p)] do
45:          check_flag ← CHECK_NEIGHBOR(new_list,arr_p[ind])
46:          if check_flag == False then
47:              new_list ← new_list + arr_p[ind]
48:          end if
49:      end for
50:      return new_list
51:  end function
52:  function COMPUTE_DIS(ulti_top)
53:      if length(ulti_top) < 2 then
54:          return 0
55:      else
56:          distance_arr ← []
57:          for each index ∈ [1,length(arr_p)] do
58:              n_p ← ulti_top[0]
59:              tep_p ← ulti_top[1]
60:              dis ← sqrt((tep_p[0]-n_p[0])^2) + (tep_p[1]-n_p[1])^2)
61:              distance_arr ← distance_arr + dis
62:          end for
63:          return distance_arr
64:      end if
65:  end function
```

### 3.2. GAN-based path prediction module

In terms of path prediction, traditional methods such as Kalman filtering have achieved good results and have been used in engineering. However, the Kalman filter is only good for tracking objects with linear motion. It is difficult to deal with the single target tracking problem with different types of motion trajectories. Especially when the target is occluded or disappears, its measured value is uncertain. At this time, it is difficult to make effective predictions.

As shown in Figure 3, in the common target tracking datasets such as VOT2018, the motion paths of different targets mostly do not follow the linear principle. In this case, it is difficult to predict the trajectory well with Kalman filter, but the Long Short Term Memory (LSTM) performs better. Because LSTM benefits from its internal mechanism, it is more suitable for solving the prediction problem of nonlinear motion trajectories. Social-GAN [30], as a representative algorithm, has excellent performance in path prediction. In order to cope with the diversified path of the target, we improved the LSTM algorithm and proposed a path prediction model. Our model takes the target center coordinates as input and uses GAN to generate diversified path samples to obtain the predicted trajectory of the target in the next few frames.

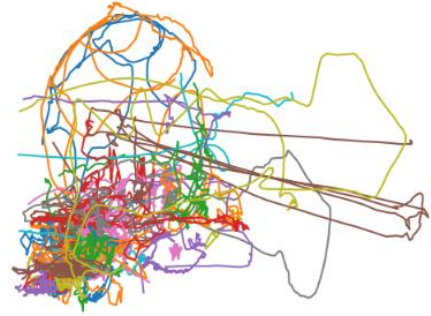

Figure 3. Motion trajectories of all sequences in VOT2018

We assume that the trajectory of the target is $\mathbb{X} = X_1, X_2, X_3, \ldots, X_n$, and the label of the predicted trajectory in the future is $\widehat{\mathbb{X}} = \widehat{X}_1, \widehat{X}_2, \widehat{X}_3, \ldots, \widehat{X}_n$. The time from $t = 1$ to $t = t_{obs}$ can be regarded as observable time, and the corresponding trajectory is an observable trajectory, which is denoted as $X_i = (x_i^t, y_i^t)$. The time from $t = t_{obs} + 1$ to $t = t_{pred}$ is recorded as the predicted time, and the corresponding trajectory is the predicted trajectory. The label is defined as $\widehat{X}_i = (\hat{x}_i^t, \hat{y}_i^t)$.

As shown in Figure 4, a GAN-based path prediction module is proposed, which includes a generator $\boldsymbol{G}$ for generating a sample of the diverse path, and a discriminator $\boldsymbol{D}$ for discriminating whether the path generated by $\boldsymbol{G}$ is in line with expectations. The generator takes $z$ as input and outputs the sample $G(z)$. The discriminator takes the sample x as an input and outputs $D(x)$. $D(x)$ indicates the probability that the sample $x$ parsed by $G(z)$ is true. The objective function is:

$$\min_G \max_D V(G,D) = \mathbb{E}_{x \sim p_{data}(x)}[log D(x)] \\ + \mathbb{E}_{z \sim p(z)}[log D(1 - D(G(z)))] \quad (4)$$

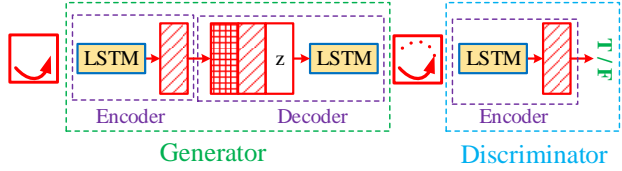

Figure 4. The GAN-based path prediction module includes a generator $G$ and a discriminator $D$. The generator $G$ is used to generate diversified path samples of the target in the next few frames, and the discriminator $D$ is used to determine whether the trajectory generated by the $G$ is reasonable.

3.3. Siamese Network with occlusion supervision

We represent an image sequence in the dataset as $X: \{Z, D_i\}_{i=2}^T$, where $Z$ is the image of template frame in the sequence, $D_i$ is the ith detection frame, and $T$ is the number of images in the sequence. In the training of the siamese network, the template frame of a certain sequence is fixed as an area containing the target in the first frame. The detection frame usually starts from the second frame of the sequence, and is usually obtained by cropping the position estimated by the coordinates of the target in the previous frame. When the network is forward-propagating, the paired template frame image and the detected frame image are sent to the network. After the output is obtained, the associated loss is calculated and back-propagating.

The pre-training of the algorithm is unsupervised, that is, it is not trained using occlusion information. Similar to a typical detection network, the siamese network is divided into a target-background classification loss and a bounding box regression loss. The classification loss is used to distinguish foreground and background, and the regression loss is used to pinpoint the location of the bounding box. From a macro perspective, the classification loss can be expressed as follows:

$$L_{cls}^{pos} = \sum_{i=2}^T y_{pos}^{(i)} \log \hat{y}_{pos}^{(i)} + (1 - y_{pos}^{(i)})\log(1 - \hat{y}_{pos}^{(i)}) \quad (5)$$

$$L_{cls}^{neg} = \sum_{i=2}^T y_{neg}^{(i)} \log \hat{y}_{neg}^{(i)} + (1 - y_{neg}^{(i)})\log(1 - \hat{y}_{neg}^{(i)}) \quad (6)$$

Where $y_{neg}^{(i)}$ indicates the predicted value when the prediction result is a negative sample and $\hat{y}_{neg}^{(i)}$ is the label of the negative sample. $y_{pos}^{(i)}$ indicates the predicted value when the prediction result is a positive sample, and $\hat{y}_{pos}^{(i)}$ is the label of the positive sample. The classification loss is:

$$L_{cls} = \lambda^{neg} L_{cls}^{neg} + \lambda^{pos} L_{cls}^{pos} \quad (7)$$

Where $\lambda^{neg}$ is the weight of the negative sample loss, and $\lambda^{pos}$ is the weight of the positive sample loss. For the regression of the bounding box, we use the $L_1$ loss function:

$$L_{reg} = \sum_{i=2}^T \left\lVert b_i - \hat{b}_i \right\rVert_1 \quad (8)$$

$b_i$ is the predicted bounding box, and $\hat{b}_i$ is the label. The total loss is:

$$L_{total} = \alpha L_{cls} + \beta L_{reg} \quad (9)$$

$\alpha$ is the weight of the classification loss, and $\beta$ is the weight of the regression loss.

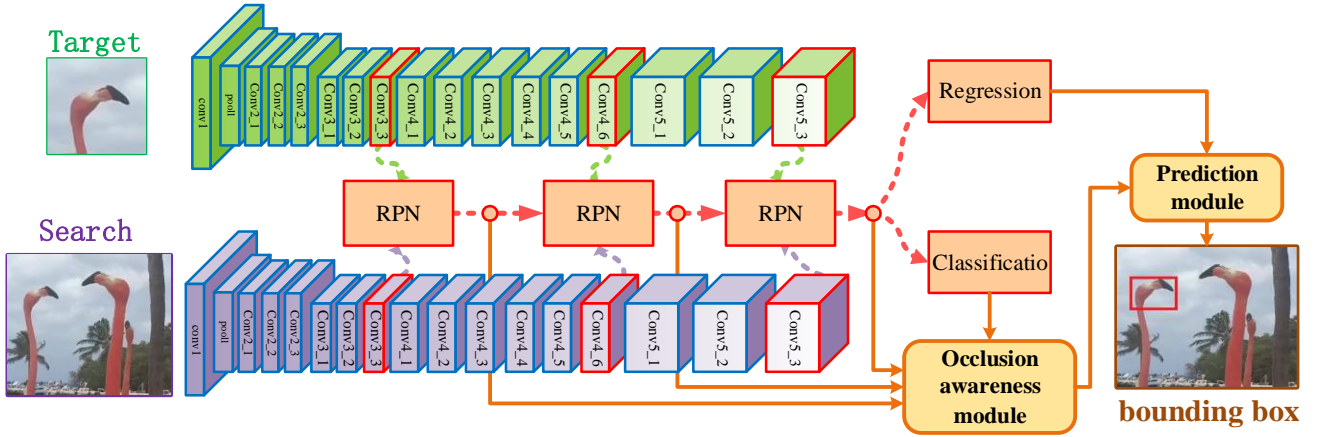

Figure 5. The structure diagram of the algorithm. The tracking target and potential interferences are obtained by the RPN, and the occlusion awareness module is used to judge whether the interference object occludes the target. If no occlusion occurs, continue tracking. If occlusion occurs, the prediction module is activated, and the motion trajectory of the target in the subsequent frames is predicted according to the motion trajectory of the target before occlusion. The predicted result is used to replace the target position feature obtained by the original tracking algorithm.

In order to better capture the target features, we added occlusion supervision information when training the data. When the sequence is occluded, we treat the sample as a negative sample for the classification branch, and the sample is still back-propagated for the regression branch. The classification loss at this time is as follows:

$$L_{cls} = \gamma \lambda^{neg} L_{cls}^{neg} + (1 - \gamma) \lambda^{pos} L_{cls}^{pos} \qquad (10)$$

$\gamma$ is an indicator for judging whether occlusion occurs. When the sequence is an occlusion sequence, $\gamma$ is 1, otherwise $\gamma$ is 0.

The training of occlusion supervision is carried out on the basis of unobstructed supervision. Our purpose is to correct the error information learned by the model from the occlusion sequence, so that the response of the output from network is closer to the real target, and it is more helpful to the occlusion awareness module to make a judgment.

So far, all the components of the Siam-GAN algorithm have been introduced. The overall structure of the algorithm is shown in Figure 5. The main idea is to obtain the image information through the siamese network and obtain the occlusion information of the target through the occlusion awareness module. If occlusion occurs, the subsequent coordinates of the target are predicted by the prediction module. Our algorithm can choose the number of predicted frames, or you can select the center coordinates of the target, the upper left corner coordinate point or the lower right corner coordinate point for prediction. Our model is not only suitable for occlusion situations in actual tests, but also produces good results for some disturbances that gradually approach the target without occlusion.

## 4. Experiment

### 4.1. Occlusion supervision training and evaluation

The backbone network of our architecture was pre-trained on ImageNet and used for image tagging. Experiments have shown that the result of such pretraining is an excellent initialization model. We trained the network on the training set of COCO [6], ImageNet DET [31], ImageNet VID and YouTube-Bounding-Boxes [32] datasets and learned how to measure the similarity between general objects in visual tracking problems. In training and testing, we used a single-scale image with a template area of 127 pixels and a search area of 255 pixels.

We mainly use the VOT2016 and VOT2018 data sets [33,34,35] for short-term single-target tracking test. In tracking, objects often appear to be occluded, which is the key issue we have to solve.

### 4.2. Network Structure and Optimization

**Train.** Our network adds anti-blocking modules to the SiamRPN++ network. In the experiment, we refer to the training method of SiamRPN++ for training and setting. We reduced the stride of ResNet-50 and connected two sibling convolution layers to it, then used 5 anchors to perform classification and bounding box regression operations. We also connected three randomly initialized 1x1 convolution layers to conv3, conv4, conv5 to reduce the feature size to 256.

Our training method uses a stochastic gradient descent (SGD). We use a synchronous SGD of 2 GPUs, with a total of 32 pairs per minibatch (16 pairs per GPU), which takes 48 hours to converge. We set the warmup learning rate to 0.001 in the first 12 hours to train the RPN branch. In the last 36 hours, the entire network was end-to-end training, and the learning rate was exponentially decayed from 0.005 to 0.0005. Set the weight decay to 0.0005 and the momentum to 0.9. The training loss is the sum of the classification loss and the regression standard smooth $L_1$ loss.

| peak distance $d$ | 3 | 3.25 | 3.5 | 3.75 | 4 | 4.5 | 5 | 5.5 | 6 |
|---|---|---|---|---|---|---|---|---|---|
| EAO | 0.381 | 0.405 | 0.384 | 0.357 | 0.36 | 0.28 | 0.316 | 0.264 | 0.234 |
| Accuracy | 0.529 | 0.576 | 0.621 | 0.492 | 0.519 | 0.486 | 0.375 | 0.403 | 0.334 |
| Robustness | 0.316 | 0.291 | 0.312 | 0.326 | 0.327 | 0.346 | 0.352 | 0.368 | 0.399 |
| score $s$ | 0.55 | 0.6 | 0.65 | 0.7 | 0.75 | 0.8 | 0.85 | 0.9 | 0.95 |
| EAO | 0.224 | 0.264 | 0.287 | 0.313 | 0.365 | 0.394 | 0.413 | 0.396 | 0.372 |
| Accuracy | 0.459 | 0.487 | 0.521 | 0.553 | 0.581 | 0.613 | 0.651 | 0.623 | 0.605 |
| Robustness | 0.405 | 0.368 | 0.359 | 0.335 | 0.331 | 0.318 | 0.296 | 0.271 | 0.289 |
| index $\varepsilon$ ($i$=0.5) | 0.55 | 0.6 | 0.65 | 0.7 | 0.75 | 0.8 | 0.85 | 0.9 | 0.95 |
| EAO | 0.305 | 0.33 | 0.346 | 0.365 | 0.364 | 0.392 | 0.419 | 0.387 | 0.366 |
| Accuracy | 0.468 | 0.509 | 0.523 | 0.561 | 0.589 | 0.622 | 0.639 | 0.638 | 0.596 |
| Robustness | 0.415 | 0.389 | 0.356 | 0.337 | 0.315 | 0.298 | 0.279 | 0.269 | 0.285 |
| weight $i$ ($\varepsilon$=0.85) | 0.1 | 0.2 | 0.3 | 0.4 | 0.5 | 0.6 | 0.7 | 0.8 | 0.9 |
| EAO | 0.287 | 0.296 | 0.328 | 0.355 | 0.376 | 0.376 | 0.389 | 0.421 | 0.403 |
| Accuracy | 0.506 | 0.539 | 0.563 | 0.58 | 0.598 | 0.615 | 0.632 | 0.67 | 0.623 |
| Robustness | 0.342 | 0.327 | 0.3 | 0.27 | 0.247 | 0.219 | 0.199 | 0.186 | 0.21 |

Table 1. Test results obtained with different indicators

**Occlusion threshold.** In order to study the impact of the timing of starting the prediction module on the tracking performance, we first trained three variants with a single RPN on ResNet-50 and then aggregated all three layers. We define three thresholds for judging occlusion: one is based on the peak distance $d$ in the heat map. If the distance $d$ between the tracked target and other peaks is less than the set peak distance $d_t$, it is judged as occlusion. The second is based on the score $s$ of each frame in the tracking process. If $s$ is less than the set minimum score $s_t$, it is judged as occlusion. The third is to define a comprehensive index $\varepsilon$ after comprehensively considering the distance and the score, and $\varepsilon$ is expressed as:

$$\varepsilon = i \times \frac{s}{0.95} + (1-i) \times \frac{d}{5.5} \qquad (11)$$

Where $i \in (0,1)$. The weight of the distance and score is adjusted by changing the value of $i$. $s$ and $d$ are first normalized and then summed. If $\varepsilon$ is less than the set threshold $\varepsilon_t$, it is judged as occlusion, and when occlusion is started, the prediction module is started. We use the VOT2016 and VOT2018 data sets for evaluation. We change the distance threshold $d_t$, the score threshold $s_t$, and the comprehensive threshold $\varepsilon_t$ including distance and score, respectively. The test results are shown in Table 1. This shows that using the method of comprehensive indicator evaluation, let $i$=0.8, $\varepsilon_t$=0.85 works best.

**Predicted length.** In theory, the more points observed in the past, the higher accuracy of the predicted future coordinates that can be obtained. However, too many points in the experiment will lead to a decrease in accuracy. In summary, we experimented with the number of observed points, training 2, 4, and 6 observation points respectively and predicting the next 2 points. The evaluation index is the average displacement error (ADE), and the data set is ETH. The calculation formula for the evaluation is:

$$ADE = 0.5 \times \sum_{i=1}^{2} \left\| Y_i - \hat{Y}_i \right\|^2 \qquad (12)$$

The result is shown in Figure 6.

It can be observed from Figure 6 that when the number of samples is between 20 and 60, the effect of the four observation points is the best. Only when the sample size is 10, 6 observation points will be better than 4 observation points. Considering comprehensively, The ADE of the four observation points is smaller than the ADE of the six observation points. Therefore, for the prediction of two consecutive frames in the future, the use of four observation points works best.

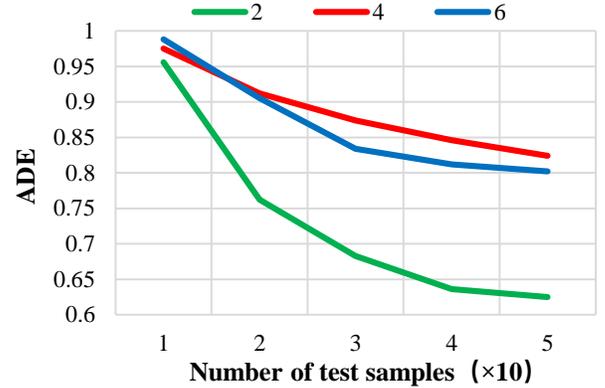

Figure 6. Different number of observation points for prediction.

|  | DaSiamRPN | SiamRPN | SA_Siam_R | SiamRPN++ | ours |
|---|---|---|---|---|---|
| EAO | 0.325 | 0.383 | 0.337 | 0.414 | 0.421 |
| Accuracy | 0.543 | 0.586 | 0.566 | 0.6 | 0.67 |
| Robustness | 0.224 | 0.276 | 0.258 | 0.234 | 0.186 |

Table 2. VOT 2018 comparison results

|  | SiamFC | SiamRPN | Staple | SRDCF | ours |
|---|---|---|---|---|---|
| EAO | 0.24 | 0.34 | 0.3 | 0.25 | 0.41 |
| Accuracy | 0.53 | 0.56 | 0.54 | 0.54 | 0.62 |
| Robustness | 0.46 | 0.26 | 0.38 | 0.42 | 0.34 |

Table 3. VOT 2016 comparison results

### 4.3. Comparison with advanced algorithms

We tested our trackers using the VOT2016 and VOT2018 data sets and compared them to 11 advanced methods such as SiamRPN++. The VOT2018 public dataset is one of the new datasets for evaluating a single object tracker for an online model, including 60 common sequences with different challenge factors. Based on the evaluation criteria of VOT2016 and VOT2018, we compare Expected Average Overlap (EAO), Accuracy (A) and Robustness (R), and non-reset based average overlap (AO) to compare different trackers. We set $i$=0.8, $\varepsilon_t$=0.85, the Accuracy-Robustness data visualization is shown in Figure 7, and the experimental results are shown in Table 2 and Table 3. Our proposed SiamGAN algorithm has made substantial progress, achieving the highest performance in EAO, accuracy and robustness standards. Compared with SiamRPN++, SiamGAN's EAO increased by 1.69%, accuracy increased by 11.67% and robustness increased by 9.3% on the VOT2018 dataset.

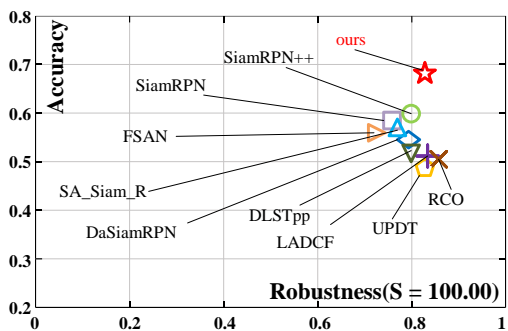

Figure 7. An Accuracy-Robustness data visualization for our algorithms and other algorithms.

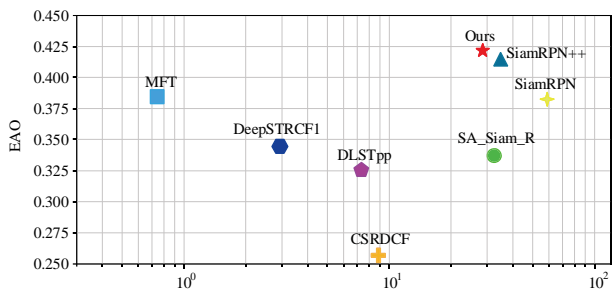

Figure 8. Speed vs EAO on VOT2018. Comparison of the quality and speed of the latest tracking methods on the VOT2018. We will visualize the Expected Average Overlap (EAO) relative to the Frames-Per-Seconds (FPS). The FPS axis is on a logarithmic scale.

We combined the speed with EAO and visualized it on the VOT2018, as shown in Figure 8. The speed was evaluated on a computer with two NVIDIA 1080ti GPUs, and other results were provided by official VOT2018 results. From the results, our algorithm achieves excellent performance, and the algorithm real-time speed (28 FPS) is slightly inferior to SiamRPN++ (35 FPS).

In actual performance, our algorithm can successfully track the target in the data set where occlusion occurs, and some results are shown in Figure 9. Our algorithm also produces good results for data sets where some interferers are getting closer to the target but no occlusion occurs.

5. Conclusions

In this paper, we propose an anti-occlusion single-target tracking algorithm called Siam-GAN, which consists of a deep siamese network, an occlusion awareness module and a path prediction module.

The backbone network of the algorithm is divided into two phases during training. The first phase has no occlusion supervision information, and the second phase incorporates occlusion supervision information. The main role is to solve the problem of using the wrong sample for the training of occlusion sequence image pairs. The actual sample should be a negative sample. The second phase fine-tuned the first phase of the network. The parameters of the occlusion awareness module are determined by the network trained in the first phase, and the parameters are determined to meet the target size of the public data. The path prediction module aims to generate a more diverse path by using the idea of Generative Adversarial Network, and is not limited to the difference between linear and nonlinear.

Using this algorithm, we have achieved excellent results on some public datasets, showing the effectiveness of the anti-occlusion module. Not only that, but we have incorporated this idea into practical applications, making the tracking algorithm more robust and able to adapt to the interference caused by occlusion.

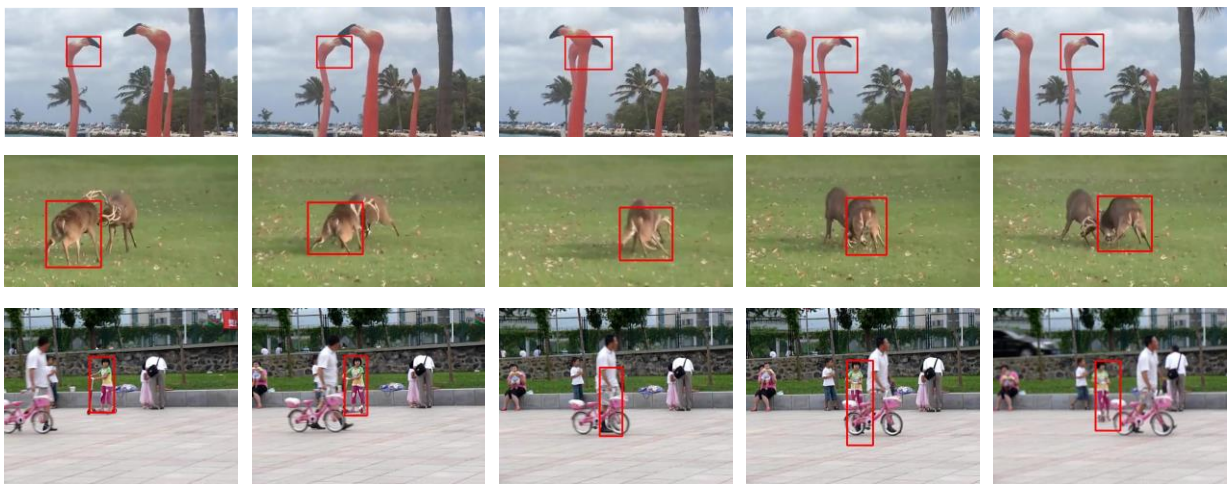

Figure 9. Tracking results of anti-occlusion algorithm